\documentclass{article}

\usepackage{arxiv}

\usepackage[utf8]{inputenc} 
\usepackage[T1]{fontenc}    
\usepackage{hyperref}       
\usepackage{url}            
\usepackage{booktabs}       
\usepackage{amsfonts}       
\usepackage{nicefrac}       
\usepackage{microtype}      
\usepackage{lipsum}
\usepackage{graphics}
\usepackage{graphicx}
\usepackage{algorithm,algorithmicx,algpseudocode}
\usepackage{enumerate}
\usepackage{times}
\usepackage{courier}
\usepackage{xcolor}
\usepackage{url}
\usepackage[square,sort,comma,numbers]{natbib}
\usepackage{framed}
\usepackage{comment}
\usepackage{chronosys}

\usepackage{graphicx}
\usepackage{graphics}
\usepackage{longtable}
\usepackage{amsfonts}
\usepackage{amsmath}
\usepackage{xcolor}
\usepackage{multirow}
\usepackage{fancyhdr}
\usepackage{caption}
\usepackage{subcaption}
\usepackage{algorithm}
\usepackage[english]{babel}
\usepackage[utf8]{inputenc}
\usepackage{algorithm,algorithmicx,algpseudocode}
\usepackage{enumerate}

\usepackage{fancyhdr}
\usepackage{caption}
\usepackage[english]{babel}
\usepackage[utf8]{inputenc}
\usepackage{algorithm,algorithmicx,algpseudocode}
\usepackage{enumerate}
\title{t}

\author{
  Md Ifraham Iqbal{\thanks{corresponding author}}
  \\Department of Data Science\\
  Friedrich-Alexander University of Erlangen\\
  Schloßplatz 4, 91054 Erlangen, Germany \\
  \texttt{miqbal161282@bscse.uiu.ac.bd} \\
  
  \And
 Md. Saddam Hossain Mukta \\
  Department of Computer Science and Engineering\\
  United International University\\
  United City, Madani Ave, Dhaka 1212, Bangladesh\\
  \texttt{saddam@cse.uiu.ac.bd} \\
  
  \AND
  Ahmed Rafi Hasan \\
  Department of Computer Science and Engineering\\
  United International University\\
  United City, Madani Ave, Dhaka 1212, Bangladesh\\
  \texttt{ahasan191131@bscse.uiu.ac.bd} \\
}
\title{A Dynamic Weighted Tabular Method for Convolutional Neural Networks}

\begin{document}
\maketitle

\begin{abstract}

\end{abstract}  
  Traditional Machine Learning (ML) models like Support Vector Machine, Random Forest, and Logistic Regression are generally preferred for classification tasks on tabular datasets. Tabular data consists of rows and columns corresponding to instances and features, respectively. Past studies indicate that traditional classifiers often produce unsatisfactory results in complex tabular datasets. Hence, researchers attempt to use the powerful Convolutional Neural Networks (CNN) for tabular datasets. Recent studies propose several techniques like SuperTML, Conditional GAN (CTGAN), and Tabular Convolution (TAC) for applying Convolutional Neural Networks (CNN) on tabular data. These models outperform the traditional classifiers and substantially improve the performance on tabular data. This study introduces a novel technique, namely, Dynamic Weighted Tabular Method (DWTM), that uses feature weights dynamically based on statistical techniques to apply CNNs on tabular datasets. The method assigns weights dynamically to each feature based on their strength of associativity to the class labels. Each data point is converted into images and fed to a CNN model. The features are allocated image canvas space based on their weights. The DWTM is an improvement on the previously mentioned methods as it dynamically implements the entire experimental setting rather than using the static configuration provided in the previous methods. Furthermore, it uses the novel idea of using feature weights to create image canvas space. In this paper, the DWTM is applied to six benchmarked tabular datasets and it achieves outstanding performance (i.e., average accuracy = 95\%) on all of them. 

\section{Introduction}
In the era of big data, data analysis has become one of the fundamental tools for extracting information over the last few decades. Due to the abundance of available data, data analysis and machine learning have become critical components for data-driven decision-making~\cite {provost2013data}. Data is stored as both structured (i.e., excel sheets, databases) and unstructured (i.e., text, email, social media, images, videos) data. Structured data is stored as tabular data, with each column containing a distinct feature and each row containing a distinct instance. Usually, traditional classifiers such as Support Vector Machine (SVM)~\cite{noble2006support}, Logistic Regression (LogReg)~\cite{hosmer2013applied} and tree-based algorithms~\cite{breiman2001random} are preferred for tabular data analysis. These models provide satisfactory results with limited amounts of data and they outperform deep learning (DL) models on tabular datasets~\cite{iman2020comparative}. 

However, the traditional classifiers tend to give a poor performance on large datasets~\cite{ahmad2018performance}. Comparatively, DL models~\cite{lecun2015deep} perform better on large datasets due to their ability to learn complex patterns amongst the data~\cite{lecun2015deep}. In most cases, CNNs~\cite{kim2017convolutional} can show outstanding performance to classify images~\cite{wu2015deep}. CNN and Recurrent Neural Networks (RNN) have become highly effective at analyzing unstructured data~\cite{mishra2020deep}. Hence, several researchers have attempted to use the benefits of CNNs for tabular data tasks. Tabular data is the most popular form of data available~\cite{sun2019supertml}. Unfortunately, CNNs cannot be directly applied to tabular data. Hence, recently, several studies have introduced techniques that make CNN viable for tabular data analysis~\cite{sharma2019deepinsight,sun2019supertml}. DeepInsight~\cite{sharma2019deepinsight} and SuperTML~\cite{sun2019supertml} use the idea of converting tabular datasets into images. In \cite{sharma2019deepinsight} non-image datasets are converted into image datasets and they are forwarded to DL models. In \cite{sun2019supertml} the authors introduce two techniques (i.e., Equal Font-SuperTML (EFTML) and Variable Font-SuperTML (VFTML)). The VFTML gives greater image space to the more relevant features. The EFTML provides equal space for each feature. Although, in theory, the VFTML is likely to produce better results than the EFTML, the method cannot perform better in practice. Further analysis of their method shows that the TML possesses a few shortcomings. The method is not the most space-efficient when converting the data points to an image. Additionally, no parameters are used to assign the canvas space for each feature in the VFTML. However, feature analysis is key for tabular data analysis. None of  ~\cite{sharma2019deepinsight, sun2019supertml} use statistical tools for feature analysis in their experiments. Furthermore, none of the previous studies focus on the applications of exploratory data analysis and their impact on the performance of CNNs. 

In this paper, the Dynamic Weighted Tabular Method (DWTM) is proposed for applying Convolutional Neural Networks (CNN) to tabular datasets. The method is the first of its kind that uses feature weights to create images for applying CNNs. The primary emphasis of this study is to create a technique that uses CNN models for tabular data analysis while prioritizing the essential features. Additionally, the designed system must be robust so it can deal with datasets of all types and sizes. In this direction, the proposed method uses statistical techniques (i.e., Pearson Correlation, Chi-Square) to compute the weights of each feature. The features are then arranged in descending order based on the calculated weights. Each feature is assigned space in the image canvas based on their corresponding weights. The features are inserted accordingly based on their weights. The algorithm for the DWTM is based on the best-fit approach to ensure maximum utilization of the image canvas space based on the feature weights. Each datapoint is converted into images and are fed into CNNs (i.e., Resnet-18~\cite{he2016deep}, DenseNet~\cite{huang2017densely} and Inception~\cite{szegedy2015going}) for analysis. To the best of our knowledge, no previous studies have proposed this approach before. In this study, the DWTM is applied on the top datasets from the UCI Machine Learning Repository~\cite{Dua:2019}. In short, the study has the following contributions:

\begin{itemize}
    \item The method effectively allocates the canvas space based on the importance of the features. 
    \item The proposed technique uses a dynamic weighted method for applying CNN models on tabular datasets.
    \item DWTM proves to be more effective than the previous techniques for applying CNNs on tabular data.
    \item CNNs outperform the traditional classifiers to classify tabular datasets and produce unbiased results.

\end{itemize}

In this study, Section 2 presents a brief description of the previous works on similar topics. Section 3 displays the methodology of this technique. Section 4 shows the experiments and results obtained using this method. Section 5 contains the discussion section, and finally, the study is concluded in Section 6.

\section{Literature Review}
In recent times, DL has become the fundamental tool for machine learning applications~\cite{ahmad2019deep}. DL is applied in a wide domain such as computer vision (CV)~\cite{voulodimos2018deep}, natural languages processing (NLP)~\cite{kamath2019deep} and speech recognition (SR)~\cite{noda2015audio}. In this section, the main ideas from previous studies related to CNNs methods for tabular data tasks are discussed. Convolutional Neural Networks (CNN) are popular due to their unparalleled success with classifying images. Architectures like the AlexNet~\cite{hinton2012imagenet}, VGG~\cite{simonyan2014very} and deep Residual Networks~\cite{he2016deep} achieve state-of-the-art performance on the ImageNet\footnote{\url{http://www.image-net.org/}} dataset. Additionally, due to CNNs success in extracting features from given vectors~\cite{young2018recent}, CNNs are now the ideal choice for Natural Language Processing (NLP) and Image classification tasks. 

Despite the success of CNNs in other fields, much previous research attempts to use CNN for tabular data analysis. However, in most cases, traditional ML models work far better than CNN models~\cite{iman2020comparative}. Hence, CNNs for tabular data remained unexplored for an extended period. However, significant development is being made in this sector in recent times. Xu et al. ~\cite{xu2019modeling} introduce the novel Conditional GAN (CTGAN) that uses mode-specific normalization and a conditional generator. Mode-specific normalization deals with multimodal and Non-Gaussian distributions, while conditional generator deals with imbalanced columns. The authors find that their model can learn better distributions than the Bayesian network-based models in their results. In another study, the DeepInsight~\cite{sharma2019deepinsight} converts non-image data to images and uses them as input for CNNs. In this method, CNNs could simultaneously work on different types of data, including tabular data. However, this method does not work well if the dataset is small, as it will create a limited number of images for input. 

Butorovic et al.~\cite{buturovic2020novel} propose a novel method then for using CNN on tabular data analysis known as Tabular Convolution (TAC). Feature vectors are transformed into kernels using the Kernel method and convolved using the base image. The authors use Resnet with TAC for classifying gene expression and found that the results using TAC are similar to the best results that ML classifiers produce. 
In 2018, Sun et al.~\cite{sun2018super} proposed the Super Characters Method. This method is used for sentiment classification, whereby texts are converted into images using two-dimensional embeddings. This idea removes the issue of adding another separate step for Word Embedding as the images are used as input to CNNs. Further investigation shows that the results obtained using this method consistently outperform the other methods for sentiment analysis. As an update based on this work, the authors introduce the SuperTML method~\cite{sun2019supertml}. The same idea of two-dimensional embeddings is used. However, this time the SuperTML method is applicable for tabular data. For each instance, a separate image is created. A different textbox is allocated in the image for each feature, where the values in each row are inserted. In this paper, the authors propose two variations of the SuperTML method, the EF and VF variations. The features are given equal importance in the EF variation, while the most important feature has the largest feature size in the VF variation. Results show that the SuperTML produces state-of-the-art performance on benchmark tabular datasets. 


Despite the recent success of DL methods in tabular data, none emphasize feature importance, which is a critical element in tabular data analysis. Previous studies predominantly use static methods, which may not work well for all tabular datasets. The method proposed in this study dynamically allocates image canvas space to the features based on their strength of association with the class label.

 \begin{figure}[ht]
    \includegraphics[scale=0.16]{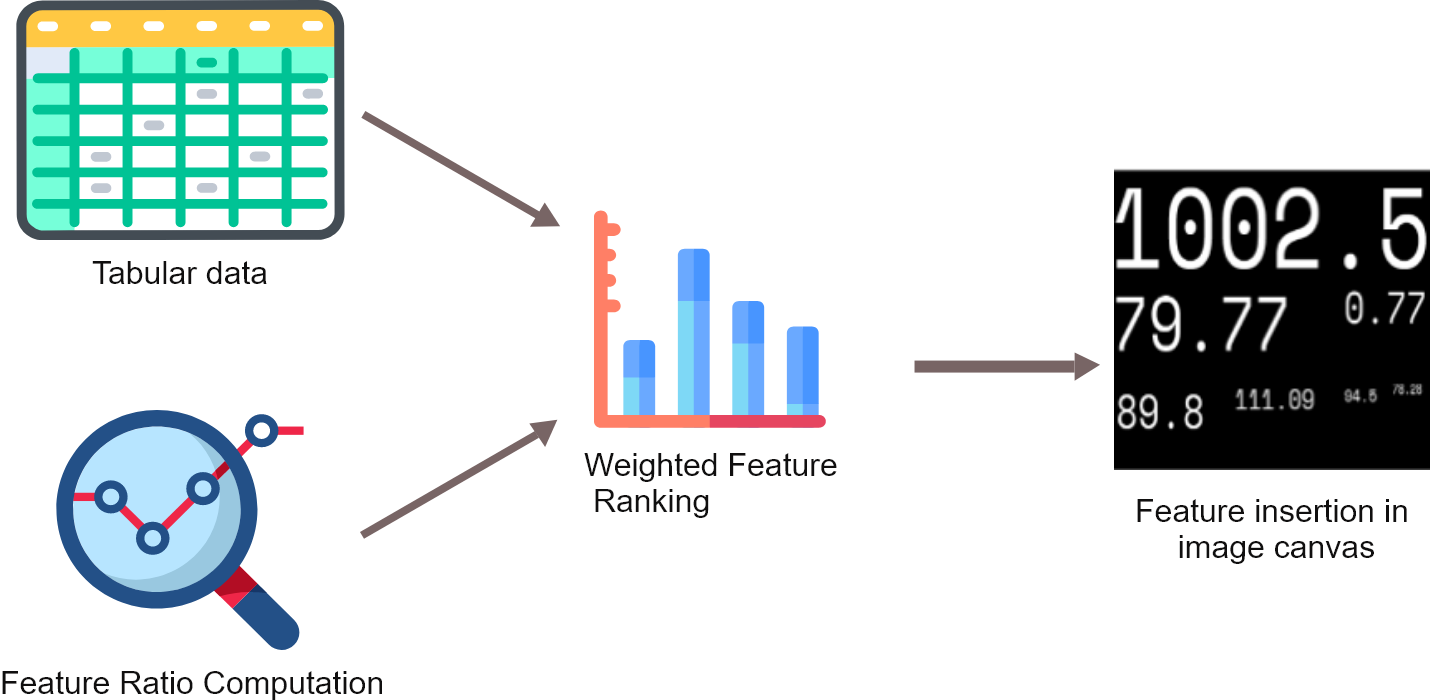}
    \caption{High level overview of DWTM.}
    \label{Process}
\end{figure}

\section{Methodology}
This section gives the methodology for the Dynamic Weighted Tabular Method (DWTM) in detail. Tabular datasets contain multiple features, where some are more associated with the class than others. In the experiments for this study, statistical techniques (i.e., Pearson Correlation, Chi-Square) are used to compute the weights of each feature. The features are then arranged in descending order based on the calculated weights. Each feature is assigned space in the image canvas based on its corresponding weights. The method requires 4 inputs: the \textit{length} and \textit{height} of the image, the \textit{r-score} and the \textit{maximum number of characters} required for each feature. 
The DWTM calculates the length, height and area required for each feature by using the ratio of the weights of each feature to the sum of the total weights of all features and distributes the image canvas space accordingly. 

The overview of the methodology is given in Figure \ref{Process}. Algorithm \hyperref[DWTMalgo]{1} represents the algorithm used for this method.
\subsection{Structured Data to Image Algorithm}

\begin{algorithmic}[1]

\item \textbf{Algorithm 1: The DWTM Method}
\item \textbf{Input:}

\label{DWTMalgo}
\item\quad\quad \textit{m} as length of the Image
\item\quad\quad \textit{n} as height of the Image
\item\quad\quad $X_t$ as $\sum$ of \textit{pearson},$r-score$ of all features.

\item\quad\quad $F$ is a vector \textit{<contains Feature>}
\item\quad\quad $F$ as Features containing $r-score$, max no of char in $F$,  area, length and height of each feature

\item \textbf{Output:}
    \item\quad\quad s contains Starting Point for each feature
    \item\quad\quad f contains Font Size for each feature\\
    
\item\textbf{Initialize:}

\item\quad\textit{\textbf{Feature Attributes:}}
\item\quad\quad $F_R$ as $r-score$ for corresponding feature
\item\quad\quad $F.X_r$ as the ratio of $r-Score$ to $X_t$ for corresponding feature
\item\quad\quad $F.X_A$ as percentage of area corresponding feature requires
\item\quad\quad $F.X_L$, $F.X_H$ as measurement of the length and height of the area feature takes
\item\quad\quad $F.C$ as max number of character required for each feature

\item\quad\textit{\textbf{Box Attributes:}}
\item\quad\quad $B$ as Feature Box size for each feature containing the length and height of each feature
\item\quad\quad $B.X_L$ as length of corresponding feature box
\item\quad\quad $B.X_H$ as height of corresponding feature box

\item\item\textbf{Procedure:} 
\For{$f \gets F$}{}\\
\quad Feature-Box-Computation \Comment{Using Algorithm 2}

\EndFor

\item $SB$ = sortLargest$(B)$    \Comment{Feature Box sorted in descending order}
\item\item\textbf{while ($SB$ != $NULL$)}
\item\item\quad Feature-Insertion-in-Image-Canvas  \Comment{Using Algorithm 3}
\item\textbf{end while}

\end{algorithmic}

\subsection{Weight Computation of the features}\label{subsec:WC}
The Pearson Correlation coefficient is used to calculate the weights of each feature by determining the associativity between each feature and the class. The Pearson Correlation technique is widely used in research for finding the associativity between attributes or variables~\cite{luo2020probabilistic}. 
The Pearson \textit{r-score} ranges from -1 to +1 and corresponds to the strength of associativity of the features with the class label. Negative value refers to negative associativity. The \textit{r-scores} of the selected features are calculated. Afterward, the weights for each feature are computed using their calculated \textit{r-scores}. However, Pearson Correlation is not applicable for associativity calculation on categorical data~\cite{tallarida1987chi}. Hence, the Chi-Square test is used when there are categorical features in the dataset. Afterward, Cramer's V is calculated to find the strength of associativity between the two variables~\cite{prematunga2012correlational}.

\begin{equation}
F_{ratio}= \frac{r_i}{\sum(r_0-r_n)}
\label{Rratio}
\end{equation}
Equation \ref{Rratio} is used to determine the weights for the features. Here, \textit{Fratio} represents the weight and \textit{r} represents the \textit{r-score} for each feature. In case of categorical variables the \textit{r-score} is replaced by Cramer's V.

\begin{algorithmic}[1]

\item \textbf{Algorithm 2: Feature Box Computation}
\item \textbf{Input:}
\label{BoxCalculation}
\item\quad\quad \textit{m} as length of the Image
\item\quad\quad \textit{n} as height of the Image
\item\quad\quad $X_t$ as $\sum$ of \textit{pearson},$r-score$ of all features.

\item\quad\quad $F$ is a vector \textit{<contains Feature>}
\item\quad\quad $F$ as Features containing $r-score$, max no of char in $F$,  area, length and height of each feature\\

\item \textbf{Output:}
    \item\quad\quad $SB$ is a vector \textit{<contains Feature Boxes sorted by their associativity to the class>}
    \item\quad\quad $SB$ as Feature Box size for each feature containing the length, height, area and position of each feature\\

\item\textbf{Procedure:} 

\item $X_t$ = $\sum$ of $F.r-score$    \Comment{Add Summation Range}
\For{$x \gets F$}{}
\item \quad \quad $F.X_r$ = $X.RS/X_t$    \Comment{Ratio of feature’s r-score to $\sum$ of r-score}
\item \quad \quad $F.X_A$ = $X_r*m*n$    \Comment{Computing Feature Box Size in Image }
\item \quad \quad $F.X_H$ = $\sqrt{(F*A)/N}$    \Comment{Computing Height of Feature Box}
\item \quad \quad $F.X_L$ = $N * F.X_H$    \Comment{Computing Length of Feature Box}
\EndFor
\item Insert $F.X_L$, $F.X_H$, $F.X_A$ in $B$ 
\item $F.X_A$ = $[\{X_L\}*\{X_H\}]$    
\item $SB$ = sortLargest$(B)$    \Comment{Feature Box sorted in descending order}

\end{algorithmic}

\subsection{Canvas Size Allocation}\label{subsec:CC}
In this subsection, the area required for each feature in the image canvas is calculated using the \textit{Fratio}. The formula for calculating the total area of each feature required in the image canvas is calculated using Equation \ref{F-area}. In this study, it is assumed the length and width of the total image are \textit{m} and \textit{n}, respectively. Afterward, the length and height required for each feature in the image are quantified. The height for each feature is the square root of the ratio of the total area and characters required for the corresponding feature and is calculated by using Equation \ref{Fheight}. Next, the length of each feature is calculated using the product of the height and character number for that feature. For this study, the monospace~\cite{rello2016effect} font type is used as this font consists of the same length and height for each character. Algorithm \hyperref[BoxCalculation]{2}for this subsection. For example, if there exist three features, with weights of 0.5, 0.3 and 0.2, and characters required 2, 3 and 4, respectively. Assuming the image is of size 128 by 128, the area of feature-1 is calculated by using Equation \ref{F-area}. The calculated area for feature 1 is 8192 units. Then the height using Equation \ref{Fheight} (64 units) and then the length using Equation \ref{Flength} (128 units) are calculated. The same process is repeated for features 2 and 3. 
After the process is complete, all the dimensions (i.e., length, height, area) for each feature are acquired. 
\begin{equation}
F_{Area}= r_i*m*n
\label{F-area} 
\end{equation}
 
\begin{equation}
F_{Area} = F_{Height}*F_{Length}
\label{Farea} 
\end{equation}

\begin{equation}
F_{Height} = \sqrt{\frac{F_{Area}}{F_{char}}}
\label{Fheight}
\end{equation}

 \begin{equation}
F_{Length} = {F_{Height}F_{char}}
\label{Flength}
\end{equation}

\subsection{Optimized Canvas Area Division}\label{subsec:OI}
After computing all the dimensions for each feature, it is now possible to insert the feature into each image. However, the points at which these features are to be inserted into the image canvas for the optimal solution are still unknown. Hence, the best fit solution to the problem is designed. Each feature is assumed to be a box of length $F_l$ and height $F_h$. The image is considered as an empty canvas of size \textit{m} and \textit{n}. At the start of the procedure, the features are arranged in descending order based on their area. Then the algorithm \hyperref[FeatureInsertion]{3} is used to insert the features into the available space in the image canvas. In this procedure, each pixel (assuming each pixel as a value in an array) in the image is iterated until the maximum possible space for each feature is found. The feature is inserted into that space. 

\begin{algorithmic}[1]

\item \textbf{Algorithm 3: Feature Insertion in Image Canvas}
\item \textbf{Input:}
\label{FeatureInsertion}
\item\quad\quad \textit{m} as length of the Image
\item\quad\quad \textit{n} as height of the Image
\item\quad\quad $SB$ is a vector \textit{<contains Sorted Feature Boxes>}
\item\quad\quad $B$ as Feature Box size for each feature containing the no. of characters, length, height, area and position of each feature\\

\item \textbf{Output:}
\item\quad\quad \textit{s} contains Starting Point of each feature
\item\quad\quad \textit{f} contains Font Size of each feature

\item\item\textbf{Procedure:}
\item\item I = [m]*[n]   \Comment{Image Space}
\For{$k \gets SB$}{}
        \Comment{Check Canvas availability for SB[k]}
                    \If {$flag$ == $1$}
                         \item\State Canvas not available \\
                         \quad\quad\quad \textbf{Continue}
                    \item \EndIf

                    \item \If {$flag$ == $0$}\\
                        \quad\quad\quad Canvas is available\\
                        \quad\quad\quad Insert feature into image canvas
                        \State pop SB[k] from SB \Comment{Feature Inserted}
                    \EndIf

\item \If {SB != empty}\\
    \quad\quad\quad Feature-Trim \Comment{Using Algorithm 4}
    \item \EndIf
\EndFor

\end{algorithmic}

\begin{figure}[h]
    \centering
    \includegraphics[scale=0.25]{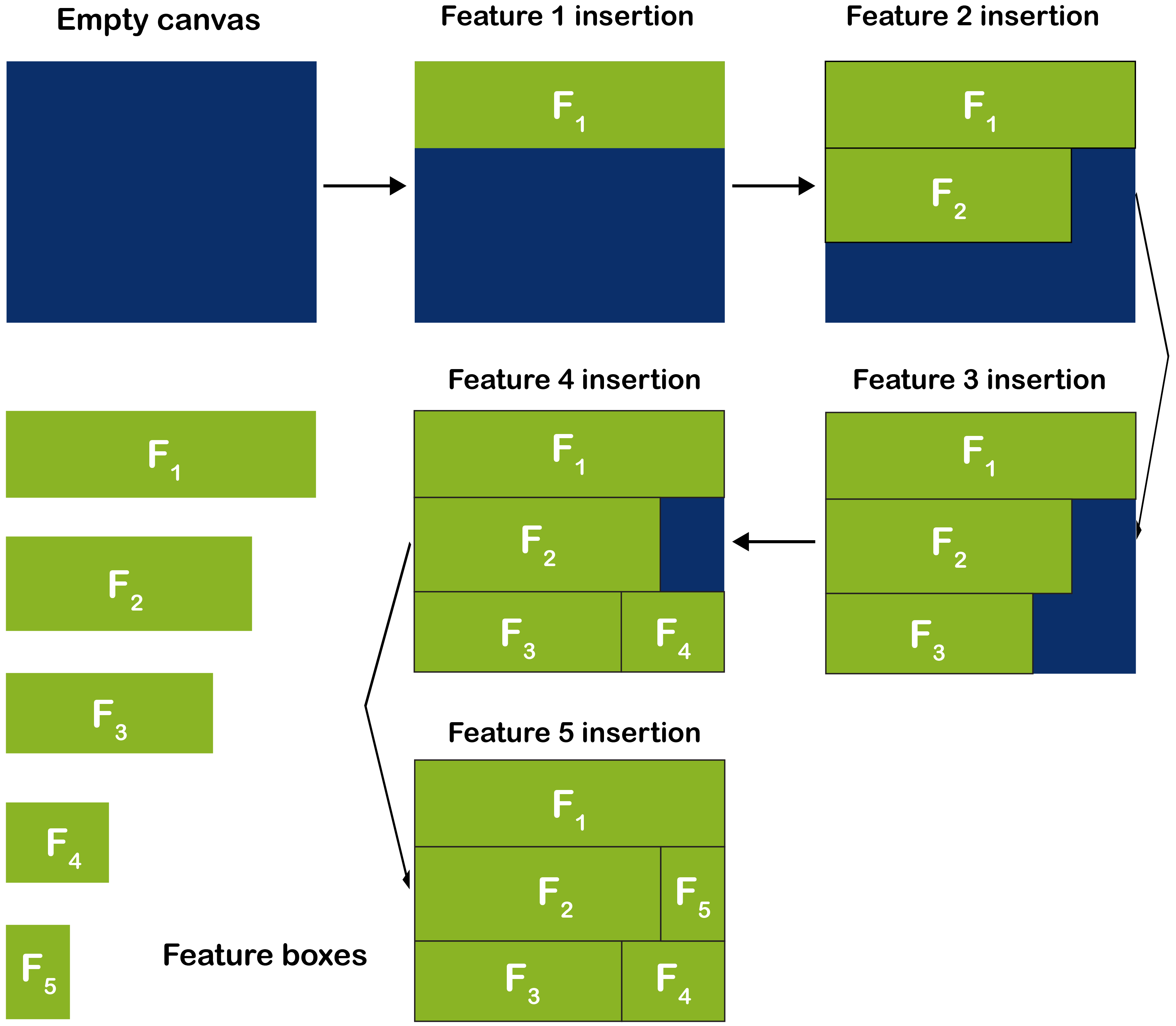}
    \caption{A sample simulation for feature insertion into canvas space using DWTM}
    \label{simulation}
\end{figure}

A sample simulation for this procedure is shown in figure \ref{simulation}. When the required space is found, it is checked if the corresponding rows and columns equalling the length and height of the feature box are empty or not. If the image space is empty, the feature is added to the canvas space and the space is marked as filled for the corresponding feature. The starting point is also stored. The process is repeated until all the features are given space in the image canvas. In the end, the dimensions and the starting points for all the features are acquired. The height of each feature is equivalent to the font size in the image.

In some cases, some of the features fail to be inserted by the algorithm. This situation occurs due to the unavailability of abundant space in one location to insert the feature box. Hence, the feature box cannot be inserted despite enough overall blank space availability in the image. The features which face this difficulty fall in the lower half of the priority list. To overcome this issue a trim feature is used. In the trim feature font size of each uninserted feature is decreased by one until all the features are inserted into the image. To do this algorithm \hyperref[trim]{4} is used which reduces the font size of the remaining features by 1. Again the Weighted Feature Insertion procedure is called. This process continues until each feature is inserted or has a font size of 0. The result is that the least important features heuristically have lesser space or are removed altogether.

\begin{algorithmic}[1]

\item \textbf{Algorithm 4: Feature Trim}
\item \textbf{Input:}
\label{trim}
\item\quad\quad $UB$ is a vector \textit{<contains uninserted Feature Boxes>}
\item\quad\quad $UB$ as Feature Box size for each feature containing the no. of characters (n), length (l), height (h), and area (a) of each feature\\

\item \textbf{Output:}
\item\quad\quad $TB$ is a vector \textit{<contains Trimmed Feature Boxes>}
\item\quad\quad $TB$ as Feature Box size for each feature containing the no. of characters (n), length (l), height (h), and area (a) of each feature

\item\item\textbf{Procedure:}
\item\For{$b \gets UB$}{}\\
    \quad\quad b.h = b.h - 1\\
    \quad\quad b.l = bl. - n\\
    \quad\quad b.a = b.l * b.h
\EndFor

\end{algorithmic}

\subsection{Image Creation}\label{subsec:IC}
\begin{figure}[ht]
    \centering
    \includegraphics[scale=0.8]{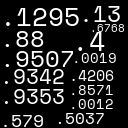}
    \caption{Sample image of a datapoint using DWTM}
    \label{sample}
\end{figure}

The algorithm returns each feature's starting point, area, length, and height. Then using OpenCV~\cite{bradski2008learning}, each feature is inserted into the image one after another into the image canvas using the information determined earlier. An image is created by using all the features of each data point. These values are rounded to the maximum character sizes that are allowed for those features and then inserted using a \textit{monospace} font, which is available in \textit{OpenCV}. Figure~\ref{sample} presents a sample image produced by this method on a classified dataset.

\subsection{The Convolutional Neural Networks used in the experiments}
The image dataset is created based on the above methodology. The dataset is then split into training and testing sets and forwarded to the DL models for training. In this case the popular Residual Network-18~\cite{he2016deep}, Densely Connected Convolution Networks~\cite{huang2017densely} and Inception~\cite{szegedy2015going} model are used. 
The study in ~\cite{huang2017densely} builds on the idea of the shortcut connections between layers and proposes a Densely Connected Convolutional Network (DenseNet). All the layers with matching feature maps are connected in this network, thus significantly boosting the information flow between the layers. Results from the study show that DenseNet performs better than other state-of-the-art architectures in terms of computational efficiency. In ~\cite{szegedy2015going} the authors proposed the GoogLeNet, also known as Inception Network. This network applies multiple kernels on the same level, thus enabling the network to tackle overfitting and reducing Deep Neural Networks' computational expense. As a result, it is feasible to use Inception in big data tasks~\cite{movshovitz2015ontological}. An updated version of the network, Inception-v3~\cite{szegedy2016rethinking}, provides state-of-the-art results for computer vision tasks while being much cheaper computationally. In this study, ReLU activation is used. ReLU has a linear scale for positive values and a value of 0 for negative instances. LR values between 0.0001 and 0.00005 and 30 epochs are used to train the models. Additionally, the Stochastic Gradient Descent (SGD), Adamax, Adam, Adadelta and Adagrad optimizers are used.

\section{Evaluation Experiments}
The DWTM is applied over selected benchmark datasets from UCI Machine Learning Repository. The results of DWTM are compared with the three popular traditional classifiers: Logistic Regression, Random Forest and Support Vector Machine. The correlated features inserted into each canvas while using DWTM are selected when applying the traditional classifiers.

\subsection{Dataset Selection}
Details of the selected benchmarked datasets used to evaluate the performance of DWTM is given below. The UCI Machine Learning Repository~\cite{Dua:2019} has an extensive collection of preprocessed and benchmarked datasets that researchers use. Hence, this repository is used to evaluate the DWTM. Previous studies show that, usually, deep learning models require a large amount of data to be trained properly~\cite{kamilaris2018deep}. However, to test the robustness of our model, datasets that contain less than 1000 instances are selected alongside large datasets. Additionally, emphasize is given on using medical datasets. It is critical in disease diagnosis for models not to be biassed towards one class~\cite{fletcher2021addressing}. As a result, the scopes of DWTM for disease diagnosis is also evaluated. The datasets selected for this study are the Cleveland dataset\footnote{\url{https://archive.ics.uci.edu/ml/datasets/heart+disease}}, the Early Stage Diabetes Risk Prediction~\cite{islam2020likelihood}, the Breast Cancer Wisconsin~\cite{wolberg1990multisurface} Dataset, the Iris Dataset\footnote{\url{https://archive.ics.uci.edu/ml/datasets/iris}}, the Wine Dataset\footnote{\url{https://archive.ics.uci.edu/ml/datasets/wine}} and the Adult Dataset\footnote{\url{https://archive.ics.uci.edu/ml/datasets/adult}}. Details of the selected datasets are given in table \ref{dataset_details}.

\begin{table}[ht]
\centering
\caption{Details of the selected datasets.}
\label{dataset_details}
\scalebox{0.9}{
\begin{tabular}{|l|l|l|l|l|}
\hline
Dataset & \# of Inst. & \# of features & \# of Classes & \% of train. inst. \\ \hline
Cleveland 
& 303 & 14 & 2 & 85\%\\ \hline
Diabetes & 520 & 17 & 2 & 80\%\\ \hline
Iris & 150 & 4 & 3 &85\%\\ \hline
Wine & 178 & 13 & 3 &85\%\\ \hline
Breast Can. & 699 & 10 & 2 & 80\%\\ \hline
Adult & 48842 & 14 & 2 & 67\%\\ \hline  
\end{tabular}
}
\end{table}

The Cleveland dataset\footnote{\url{https://archive.ics.uci.edu/ml/datasets/heart+disease}} consists of data from 303 patients correlated to the patient's risk of heart disease. The dataset contains 76 features. However, the website states that most researchers preferred to use a subset of 14 features from this dataset. Hence, these 14 features from the dataset for the experiments in this study. The class column, \textit{condition} contains values from zero to five, which indicates the patient's risk of heart disease. The values are converted from 1-4 to one to identify the presence of heart disease in a patient. The Early Stage Diabetes Risk Prediction~\cite{islam2020likelihood} dataset was conducted in the Sylhet Diabetes Hospital in Bangladesh. To create the dataset, doctors used a questionnaire on 520 patients. Eventually, 17 features are recorded using the tests and questionnaires. Afterward, patients are tested to check for diabetes, and the results of this test represent the class for each patient. 

The Breast Cancer Wisconsin~\cite{wolberg1990multisurface} Dataset contains data of 699 patients. The data is collected using a Fine Needle Aspirate of a breast mass. There are ten features in the dataset. Class values two and four indicate that breast cancer is malignant or benign. The Iris Dataset is the most popular dataset for pattern recognition. This dataset has the most hits in the UCI Repository. The dataset consists of data from 150 iris plants. 4 features are recorded for each instance. Three classes of iris plants, each having 50 instances, are available in this dataset. The Wine Dataset contains data from the chemical analysis of wine grown in a region in Italy. The dataset contains 13 features and 178 instances and is the third most popular dataset in the UCI repository. Similar to the Iris dataset, the wine dataset consists of three classes. The Adult dataset\footnote{\url{https://archive.ics.uci.edu/ml/datasets/adult}} is the second most popular dataset in the UCI Repository. The datasets contain information of 48842 individuals. The class column is binary, with zero representing individuals who earn less than \$50K per year. The Adult dataset is the largest dataset used for these evaluation experiments. 

\subsection{Results}
\begin{table}[ht]
\caption{Result comparison of the CNN Models and Traditional Classifiers}
\label{Benchmarking}
\scalebox{1.2}{
\begin{tabular}{|l|l|l|l|l|l|l|}

\hline
Dataset & Model & Acc. & Sns. & Spc. & Prec. & F1.\\ \hline

\multirow{6}{*}{Cleveland}
&DWTM+R & 100 & 100 & 100 & 100 & 100\\ \cline{2-7}
&DWTM+D & 100 & 100 & 100 & 100 & 100\\ \cline{2-7}
&DWTM+I & 100 & 100 & 100 & 100 & 100\\ \cline{2-7}
&LogReg & 83.33 & 78.57 & 87.50 & 83.48 & 83.16\\ \cline{2-7}
&RF & 83.33 & 71.42 & 93.75 & 84.93 & 82.86\\ \cline{2-7}
&SVM & 66.66 & 42.86 & 87.50 & 69.31 & 64.11\\ \cline{2-7} \hline
\multirow{6}{*}{Diabetes}
&DWTM+R & 100 & 100 & 100 & 100 & 100 \\ \cline{2-7}
&DWTM+D & 100 & 100 & 100 & 100 & 100 \\ \cline{2-7}
&DWTM+I & 100 & 100 & 100 & 100  & 100 \\ \cline{2-7}
&LogReg & 82.69 & 82.76 & 82.61 & 82.44 & 82.53 \\ \cline{2-7}
&RF & 95.19 & 98.48 & 89.47 & 95.67 & 94.72 \\ \cline{2-7}
&SVM & 94.23 & 96.55 & 91.30 & 94.39 & 94.12 \\ \cline{2-7} \hline

\multirow{6}{*}{Breast Cancer}
&DWTM+R & 98.58 & 98.91 & 97.96 & 98.91 & 98.91 \\ \cline{2-7}
&DWTM+D & 100 & 100 & 100 & 100 & 100 \\ \cline{2-7}
&DWTM+I & 100 & 100 & 100 & 100 & 100 \\ \cline{2-7}
&LogReg & 97.14 & 95.83 & 97.82 & 96.83 & 96.83 \\ \cline{2-7}
&RF & 94.29 & 87.50 & 97.82 & 94.60 & 93.52 \\ \cline{2-7}
&SVM & 95.71 & 91.67 & 97.83 & 95.69 & 94.7 \\ \cline{2-7} \hline

\multirow{6}{*}{Iris}
&DWTM+R & 93.33 & - & - & 93.33 & 93.33\\ \cline{2-7}
&DWTM+D & 100 & - & - & 100 & 100\\ \cline{2-7}
&DWTM+I & 100 & - & - & 100 & 100\\ \cline{2-7}
&TML ~\cite{sun2019supertml} & 93.33 & - & - & - & -\\ \cline{2-7}
&LogReg & 100 & - & - & 100 & 100\\ \cline{2-7}
&RF & 100 & - & - & 100 & 100\\ \cline{2-7}
&SVM & 100 & - & - & 100 & 100\\ \cline{2-7} \hline

\multirow{6}{*}{Wine}
&DWTM+R & 100 & - & - & 100 & 100\\ \cline{2-7}
&DWTM+D & 100 & - & - & 100 & 100\\ \cline{2-7}
&DWTM+I & 100 & - & - & 100 & 100\\ \cline{2-7}
&TML~\cite{sun2019supertml} & 97.30 & - & - & - & -\\ \cline{2-7}
&LogReg & 96.30 & - & - & 97.44 & 95.64 \\ \cline{2-7}
&RF & 94.44 & - & - & 93.71 & 94.46 \\ \cline{2-7}
&SVM & 96.30 & - & - & 95.38& 94.44 \\ \cline{2-7} \hline

\multirow{6}{*}{Adult}
&DWTM+R & 100 & 100 & 100 & 100 & 100\\ \cline{2-7}
&DWTM+D & 100 & 100 & 100 & 100 & 100\\ \cline{2-7}
&DWTM+I & 100 & 1009 & 100 & 100 & 100 \\ \cline{2-7}
&TML ~\cite{sun2019supertml} & 87.60 & - & - & - & -\\ \cline{2-7}
&LogReg & 80.55 & 26.59 & 97.14 & 77.63 & 61.87\\ \cline{2-7}
&RF &  84.85 & 61.70 & 91.97 & 79.45 & 77.99\\ \cline{2-7}
&SVM & 85.30 & 57.74 & 93.77 & 80.93 & 75.76\\ \cline{2-7} \hline
\end{tabular}
}
\end{table}

The results from the experiments are displayed in Table \ref{Benchmarking}. DWTM+R, DWTM+D, and DWTM+I refer to the use of the ResNet-18, InceptionV1 and DenseNet models, respectively.
From Table \ref{Benchmarking}, it can be observed that for the Cleveland dataset, the DWTM provides significantly better results than the traditional classifiers. The DenseNet, Inception and ResNet-18 produce 100\% accuracy scores. The results show that the DWTM is a viable option for disease diagnosis due to its ability to be unbiased. The Cleveland dataset has only 303 instances. The DWTM assists the CNN models in providing better results than the traditional classifiers on the Cleveland dataset. The method again proves successful in disease diagnosis on the Diabetes dataset. The CNN models provide balanced results, which the traditional classifiers fail to deliver. Results on the Diabetes dataset provide further evidence of the model's ability to provide state-of-the-art results on a medical dataset. The DWTM shows further robustness on the Breast Cancer dataset as each CNN model produces almost the maximum possible results using the proposed method.

The Iris dataset is the most popular dataset from UCI. However, it is also the smallest dataset containing only 150 instances and four features. Furthermore, the Iris dataset has three classes of Iris plants. Despite the minimal information available, the DWTM manages to achieve maximum performance on the dataset, thus matching the performance of traditional classifiers. The Wine Dataset contains only 178 instances and three classes similar to the Iris dataset. Table \ref{Benchmarking} shows the results produced on the Wine dataset from our experiments. It is observed that the CNNs consistently produce results that are usually better or equivalent to the results from the traditional classifiers. The results of the Wine and Iris dataset also show the DWTM's ability to deal with multiclass tabular datasets. The results of the experiments on the Adult dataset are shown in Table \ref{Benchmarking}. The Adult dataset contains over 48000 instances, the largest dataset used for evaluation in this study. The performance of the DWTM models is far better than those of the traditional classifiers. The performance on this dataset shows that the DWTM can be applied to datasets of any dimension.

\section{Discussion}
The results from these experiments have a significant impact. Previous studies show that CNN models are not very effective with tabular datasets. However, results from this study demonstrate that the CNN models consistently outperform the traditional classifiers on these tabular datasets. Furthermore, in the past, it was assumed that DL only performed well with large amounts of data~\cite{kamilaris2018deep}. The results contradict this statement and provide further evidence of the usability of CNN models for small datasets.
The DWTM is also a viable option for classification tasks on medical datasets. The method effectively deals with bias much better than the traditional classifiers. The model outperforms the traditional classifiers in the Cleveland, Diabetes, and Breast Cancer datasets. Notably, in the Cleveland dataset, the traditional classifiers all produced Sensitivity scores below 0.80. On the other hand, the CNN models produce high Sensitivity scores while also producing high Specificity scores. The DWTM also shows it can perform multiclass tasks with ease. The method provides outstanding results in both the Iris and Wine datasets. The results produced are better (or similar) than those provided by the SuperTML and traditional classifiers. The Adult dataset is the only dataset used with a vast number of instances. The DWTM surpasses the performance of the traditional classifiers and the performance of the SuperTML by a considerable margin. All these results show the robustness of the model as it can be used in small, large and multiclass datasets. Furthermore, the results also indicate that DWTM makes the CNN models much more beneficial than the traditional classifiers for tabular data tasks. 
From these experiments, it can also be noted that the Adamax and Adam optimizers consistently produce better results, followed by the Stochastic Gradient Descent (SGD) with Exponential Learning Rate Decay. Learning rates between 0.0005 and 0.001 are. A learning rate of 0.001 tends to reach the desired outcomes more often.

Although previous studies of SuperTML~\cite{sun2019supertml}  and DeepInsight~\cite{sharma2019deepinsight} use the idea of converting non-image datasets into image datasets for deep learning applications, none of the studies utilize feature weights for classification. The results show that the DWTM provides similar or better outcomes when compared to the SuperTML. The DWTM is also more dynamic and robust compared to the SuperTML. It uses a novel approach by using the feature weights to create images. Compared to the DeepInsight technique, it also performs remarkably on small datasets. The DWTM provides the most significant space to the most important features based on the assigned weights. As a result, the CNN models are likely to learn the more complex patterns from the essential features. To the best of our knowledge, no previous study used feature weights for CNN models. 

This study provides the foundation of feature analysis for DL models. In this study, Pearson Score Coefficient is used for assigning weights. Future studies can use other statistical techniques and evaluate which options are the best for calculating feature weights. The Pearson Correlation and Cramer's V are preferred for these experiments as they can find the strength of associativity of a feature to the class. Alternatively, future researchers can use Fishers Correlation~\cite{asuero2006correlation} which uses a transformation technique to deal with highly correlated features. This method works much better than Pearson Correlation with highly correlated features and is another alternative for calculating feature weights. Another popular statistical technique is the Analysis of Variance (ANOVA)~\cite{kaufmann2007analysis}. This method works better when the class contains three or more levels. Hence, ANOVA is usually preferred over Pearson Correlation for multiclass datasets and regression tasks. ANOVA can also be used with DWTM for assigning weights to determine the importance of the features. Due to the introduction of DWTM, feature analysis may become key for CNN applications on tabular data tasks. Combining various feature analysis techniques and using them with the DWTM has the potential to produce state-of-the-art results in all kinds of tabular data tasks. 

In the experiments, a maximum number of 30 epochs are used. As the pre-trained CNN models are and thus, 30 epochs are sufficient in most cases. Increasing the number of epochs may produce even better results. Additionally, models like the Resnet-152, Inception-v4 and other CNN models show substantial learning capabilities. Using these models with the DWTM can further increase the performance of CNNs for tabular data. Feature Selection is a vital part of tabular data analysis. In the future, the DWTM can be upgraded further to produce the best subset of features for CNN applications on tabular datasets.   

\section{Conclusion}
This paper introduces the Dynamic Weighted Tabular Method (DWTM). The method uses feature importance with CNNs for tabular data tasks. Three CNN models are applied with the DWTM and the results are compared with the results of the three traditional classifiers. The results demonstrate that the method usually outperforms the results of traditional classifiers and the previously mentioned CNN-based methods on the six benchmarked datasets. Additionally, the method is robust for utilization in various types of datasets (multiclass, large/small). This is the first study that uses feature strength significance on tabular data for CNN models to the best of our knowledge. This study can make feature analysis vital when applying CNNs for tabular data analysis.

\section{Online Resources}
\subsection{DWTM Package Materials}

Github Repository - 
\textcolor{blue}{\url{https://github.com/AnonymousCIKM1/DWTM}}

Alternatively the drive link given below can also be used to apply the DWTM.  The link given below contains all the files for applying the method introduced in this manuscript. Please carefully read the \textbf{readme.txt} file and follow the steps. 
Drive Link - \textcolor{blue}{\url{https://drive.google.com/drive/folders/1R5IkgeyBDNTVckuyEOy04gjihVd0TNgp?usp=sharing}}
    
\subsection{Conducted Experiments for the research}
The link below contains the codes for the experiments used on the 6 datasets from UCI Repository - \textcolor{blue}{\url{https://drive.google.com/drive/folders/1cCANjPTfB20AIBd94bLbqFMlart0C4UG?usp=sharing}}

\bibliographystyle{unsrt}
\bibliography{references.bib}

\begin{thebibliography}{10}

\bibitem{provost2013data}
Foster Provost and Tom Fawcett.
\newblock Data science and its relationship to big data and data-driven
  decision making.
\newblock {\em Big data}, 1(1):51--59, 2013.

\bibitem{noble2006support}
William~S Noble.
\newblock What is a support vector machine?
\newblock {\em Nature biotechnology}, 24(12):1565--1567, 2006.

\bibitem{hosmer2013applied}
David~W Hosmer~Jr, Stanley Lemeshow, and Rodney~X Sturdivant.
\newblock {\em Applied logistic regression}, volume 398.
\newblock John Wiley \& Sons, 2013.

\bibitem{breiman2001random}
Leo Breiman.
\newblock Random forests.
\newblock {\em Machine learning}, 45(1):5--32, 2001.

\bibitem{iman2020comparative}
Mohammadreza Iman, Amy Giuntini, Hamid~Reza Arabnia, and Khaled Rasheed.
\newblock A comparative study of machine learning models for tabular data
  through challenge of monitoring parkinson's disease progression using voice
  recordings.
\newblock {\em arXiv preprint arXiv:2005.14257}, 2020.

\bibitem{ahmad2018performance}
Iftikhar Ahmad, Mohammad Basheri, Muhammad~Javed Iqbal, and Aneel Rahim.
\newblock Performance comparison of support vector machine, random forest, and
  extreme learning machine for intrusion detection.
\newblock {\em IEEE access}, 6:33789--33795, 2018.

\bibitem{lecun2015deep}
Yann LeCun, Yoshua Bengio, and Geoffrey Hinton.
\newblock Deep learning.
\newblock {\em nature}, 521(7553):436--444, 2015.

\bibitem{kim2017convolutional}
Phil Kim.
\newblock Convolutional neural network.
\newblock In {\em MATLAB deep learning}, pages 121--147. Springer, 2017.

\bibitem{wu2015deep}
Ren Wu, Shengen Yan, Yi~Shan, Qingqing Dang, and Gang Sun.
\newblock Deep image: Scaling up image recognition.
\newblock {\em arXiv preprint arXiv:1501.02876}, 7(8), 2015.

\bibitem{mishra2020deep}
Debasmita Mishra, Bighnaraj Naik, Ronali~Madhusmita Sahoo, and Janmenjoy Nayak.
\newblock Deep recurrent neural network (deep-rnn) for classification of
  nonlinear data.
\newblock In {\em Computational Intelligence in Pattern Recognition}, pages
  207--215. Springer, 2020.

\bibitem{sun2019supertml}
Baohua Sun, Lin Yang, Wenhan Zhang, Michael Lin, Patrick Dong, Charles Young,
  and Jason Dong.
\newblock Supertml: Two-dimensional word embedding for the precognition on
  structured tabular data.
\newblock In {\em Proceedings of the IEEE Conference on Computer Vision and
  Pattern Recognition Workshops}, pages 0--0, 2019.

\bibitem{sharma2019deepinsight}
Alok Sharma, Edwin Vans, Daichi Shigemizu, Keith~A Boroevich, and Tatsuhiko
  Tsunoda.
\newblock Deepinsight: A methodology to transform a non-image data to an image
  for convolution neural network architecture.
\newblock {\em Scientific reports}, 9(1):1--7, 2019.

\bibitem{he2016deep}
Kaiming He, Xiangyu Zhang, Shaoqing Ren, and Jian Sun.
\newblock Deep residual learning for image recognition.
\newblock In {\em Proceedings of the IEEE conference on computer vision and
  pattern recognition}, pages 770--778, 2016.

\bibitem{huang2017densely}
Gao Huang, Zhuang Liu, Laurens Van Der~Maaten, and Kilian~Q Weinberger.
\newblock Densely connected convolutional networks.
\newblock In {\em Proceedings of the IEEE conference on computer vision and
  pattern recognition}, pages 4700--4708, 2017.

\bibitem{szegedy2015going}
Christian Szegedy, Wei Liu, Yangqing Jia, Pierre Sermanet, Scott Reed, Dragomir
  Anguelov, Dumitru Erhan, Vincent Vanhoucke, and Andrew Rabinovich.
\newblock Going deeper with convolutions.
\newblock In {\em Proceedings of the IEEE conference on computer vision and
  pattern recognition}, pages 1--9, 2015.

\bibitem{Dua:2019}
Dheeru Dua and Casey Graff.
\newblock {UCI} machine learning repository, 2017.

\bibitem{ahmad2019deep}
Jamil Ahmad, Haleem Farman, and Zahoor Jan.
\newblock Deep learning methods and applications.
\newblock In {\em Deep Learning: Convergence to Big Data Analytics}, pages
  31--42. Springer, 2019.

\bibitem{voulodimos2018deep}
Athanasios Voulodimos, Nikolaos Doulamis, Anastasios Doulamis, and Eftychios
  Protopapadakis.
\newblock Deep learning for computer vision: A brief review.
\newblock {\em Computational intelligence and neuroscience}, 2018, 2018.

\bibitem{kamath2019deep}
Uday Kamath, John Liu, and James Whitaker.
\newblock {\em Deep learning for NLP and speech recognition}, volume~84.
\newblock Springer, 2019.

\bibitem{noda2015audio}
Kuniaki Noda, Yuki Yamaguchi, Kazuhiro Nakadai, Hiroshi~G Okuno, and Tetsuya
  Ogata.
\newblock Audio-visual speech recognition using deep learning.
\newblock {\em Applied Intelligence}, 42(4):722--737, 2015.

\bibitem{hinton2012imagenet}
Geoffrey~E Hinton, Alex Krizhevsky, and Ilya Sutskever.
\newblock Imagenet classification with deep convolutional neural networks.
\newblock {\em Advances in neural information processing systems},
  25:1106--1114, 2012.

\bibitem{simonyan2014very}
Karen Simonyan and Andrew Zisserman.
\newblock Very deep convolutional networks for large-scale image recognition.
\newblock {\em arXiv preprint arXiv:1409.1556}, 2014.

\bibitem{young2018recent}
Tom Young, Devamanyu Hazarika, Soujanya Poria, and Erik Cambria.
\newblock Recent trends in deep learning based natural language processing.
\newblock {\em ieee Computational intelligenCe magazine}, 13(3):55--75, 2018.

\bibitem{xu2019modeling}
Lei Xu, Maria Skoularidou, Alfredo Cuesta-Infante, and Kalyan Veeramachaneni.
\newblock Modeling tabular data using conditional gan.
\newblock In {\em Advances in Neural Information Processing Systems}, pages
  7335--7345, 2019.

\bibitem{buturovic2020novel}
Ljubomir Buturovic and Dejan Miljkovic.
\newblock A novel method for classification of tabular data using convolutional
  neural networks.
\newblock {\em BioRxiv}, 2020.

\bibitem{sun2018super}
Baohua Sun, Lin Yang, Patrick Dong, Wenhan Zhang, Jason Dong, and Charles
  Young.
\newblock Super characters: A conversion from sentiment classification to image
  classification.
\newblock {\em arXiv preprint arXiv:1810.07653}, 2018.

\bibitem{luo2020probabilistic}
Dandan Luo, Shouzhen Zeng, and Ji~Chen.
\newblock A probabilistic linguistic multiple attribute decision making based
  on a new correlation coefficient method and its application in hospital
  assessment.
\newblock {\em Mathematics}, 8(3):340, 2020.

\bibitem{tallarida1987chi}
Ronald~J Tallarida and Rodney~B Murray.
\newblock Chi-square test.
\newblock In {\em Manual of pharmacologic calculations}, pages 140--142.
  Springer, 1987.

\bibitem{prematunga2012correlational}
Roshani~K Prematunga.
\newblock Correlational analysis.
\newblock {\em Australian Critical Care}, 25(3):195--199, 2012.

\bibitem{rello2016effect}
Luz Rello and Ricardo Baeza-Yates.
\newblock The effect of font type on screen readability by people with
  dyslexia.
\newblock {\em ACM Transactions on Accessible Computing (TACCESS)}, 8(4):1--33,
  2016.

\bibitem{bradski2008learning}
Gary Bradski and Adrian Kaehler.
\newblock {\em Learning OpenCV: Computer vision with the OpenCV library}.
\newblock " O'Reilly Media, Inc.", 2008.

\bibitem{movshovitz2015ontological}
Yair Movshovitz-Attias, Qian Yu, Martin~C Stumpe, Vinay Shet, Sacha Arnoud, and
  Liron Yatziv.
\newblock Ontological supervision for fine grained classification of street
  view storefronts.
\newblock In {\em Proceedings of the IEEE Conference on Computer Vision and
  Pattern Recognition}, pages 1693--1702, 2015.

\bibitem{szegedy2016rethinking}
Christian Szegedy, Vincent Vanhoucke, Sergey Ioffe, Jon Shlens, and Zbigniew
  Wojna.
\newblock Rethinking the inception architecture for computer vision.
\newblock In {\em Proceedings of the IEEE conference on computer vision and
  pattern recognition}, pages 2818--2826, 2016.

\bibitem{kamilaris2018deep}
Andreas Kamilaris and Francesc~X Prenafeta-Bold{\'u}.
\newblock Deep learning in agriculture: A survey.
\newblock {\em Computers and electronics in agriculture}, 147:70--90, 2018.

\bibitem{fletcher2021addressing}
Richard~Rib{\'o}n Fletcher, Audace Nakeshimana, and Olusubomi Olubeko.
\newblock Addressing fairness, bias, and appropriate use of artificial
  intelligence and machine learning in global health.
\newblock {\em Frontiers in Artificial Intelligence}, 3:116, 2021.

\bibitem{islam2020likelihood}
MM~Faniqul Islam, Rahatara Ferdousi, Sadikur Rahman, and Humayra~Yasmin Bushra.
\newblock Likelihood prediction of diabetes at early stage using data mining
  techniques.
\newblock In {\em Computer Vision and Machine Intelligence in Medical Image
  Analysis}, pages 113--125. Springer, 2020.

\bibitem{wolberg1990multisurface}
William~H Wolberg and Olvi~L Mangasarian.
\newblock Multisurface method of pattern separation for medical diagnosis
  applied to breast cytology.
\newblock {\em Proceedings of the national academy of sciences},
  87(23):9193--9196, 1990.

\bibitem{asuero2006correlation}
Agustin~Garcia Asuero, Ana Sayago, and AG~Gonzalez.
\newblock The correlation coefficient: An overview.
\newblock {\em Critical reviews in analytical chemistry}, 36(1):41--59, 2006.

\bibitem{kaufmann2007analysis}
J{\"o}rg Kaufmann and AG~Schering.
\newblock Analysis of variance anova.
\newblock {\em Wiley Encyclopedia of Clinical Trials}, 2007.

\end{thebibliography}

\end{document}